\documentclass[letterpaper,10pt, conference]{ieeeconf}  
\usepackage{amsmath,amsfonts}
\usepackage{algorithm}
\usepackage{array}
\usepackage[caption=false,font=normalsize,labelfont=sf,textfont=sf]{subfig}
\usepackage{textcomp}
\usepackage{stfloats}
\usepackage{url}
\usepackage{verbatim}
\usepackage{times}
\usepackage{multicol}
\usepackage[bookmarks=true]{hyperref}
\usepackage{graphicx}
\usepackage[noend]{algpseudocode}
\usepackage{amsmath}
\usepackage{amssymb}
\usepackage{color}
\usepackage{gensymb}
\usepackage[noadjust]{cite}

\IEEEoverridecommandlockouts                              

\begin{document}
\title{\LARGE \bf
An Open-Source, Reproducible Tensegrity Robot\\ that can Navigate Among Obstacles
}


\author{
William R. Johnson III$^{1,Y}$,
Patrick Meng$^{1,R}$,
Nelson Chen$^R$,
Luca Cimatti$^Y$,\\
Augustin Vercoutere$^Y$,
Mridul Aanjaneya$^R$,
Rebecca Kramer-Bottiglio$^Y$,
Kostas E. Bekris$^R$\\
\thanks{$^1$Co-first authors;  $^Y$Mech. Eng., Yale Univ.;
 $^R$Comp. Sc., Rutgers Univ.}
\vspace{-12mm}
}


\maketitle

\thispagestyle{empty}
\pagestyle{empty}

\begin{abstract}

Tensegrity robots, composed of rigid struts and elastic tendons, provide impact resistance, low mass, and adaptability to unstructured terrain. Their compliance and complex, coupled dynamics, however, present modeling and control challenges, hindering path planning and obstacle avoidance. This paper presents a complete, open-source, and reproducible system that enables navigation for a 3-bar tensegrity robot. The system comprises: (i) an inexpensive, open-source hardware design, and (ii) an integrated, open-source software stack for physics-based modeling, system identification, state estimation, path planning, and control. All hardware and software are publicly available at \href{https://sites.google.com/view/tensegrity-navigation/}{https://sites.google.com/view/tensegrity-navigation/}. The proposed system tracks the robot's pose and executes collision-free paths to a specified goal among known obstacle locations. System robustness is demonstrated through experiments involving unmodeled environmental challenges, including a vertical drop, an incline, and granular media, culminating in an outdoor field demonstration. To validate reproducibility, experiments were conducted using robot instances at two different laboratories. This work provides the robotics community with a complete navigation system for a compliant, impact-resistant, and shape-morphing robot. This system is intended to serve as a springboard for advancing the navigation capabilities of other unconventional robotic platforms.

\end{abstract}


\section{Introduction}

Navigation over unpredictable terrain with obstacles remains a challenge in robotics. Tensegrity robots made with rigid struts and elastic tendons are a next-generation mobile platform promising due to their durability, adaptability, low mass, and low cost~\cite{shah2022tensegrity}. Their significant deformations and coupled dynamics, however, create formidable modeling and control challenges, which have hindered the demonstration of a complete solution for tensegrity robot navigation.

\begin{figure}[t]
    \centering
    \includegraphics[width=\linewidth]{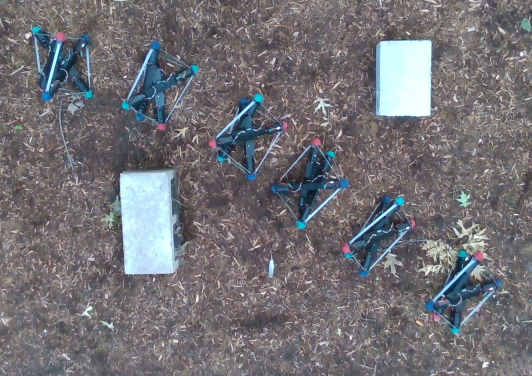}
    \vspace{-.3in}
    \caption{The open-source tensegrity robot moving around obstacles outdoors starts from the top left and autonomously navigates to the bottom right.}
    \vspace{-.25in}
    \label{fig:splash}
\end{figure}

This paper presents a complete pipeline for navigation and obstacle avoidance using a 3-bar tensegrity robot. It integrates methods in tensegrity design, modeling, and state estimation with path planning to achieve robust navigation around obstacles amidst environmental disturbances. The pipeline begins with system identification and the modeling of motion primitives via a differentiable physics engine. An open-loop planner uses these primitives to generate an action sequence to move the robot from its current pose to the goal, avoiding known obstacles. State estimation from an overhead camera closes the loop, enabling to re-plan the path after each action to correct for discrepancies between the model's prediction and the robot's state. Navigation experiments demonstrate the robot's robustness to unmodeled disturbances, including vertical drops, inclines, and granular media. The system's effectiveness is further validated through operation in an outdoor field environment (Fig.~\ref{fig:splash}).

This work also introduces a new, open-source tensegrity robot design. The hardware and software are publicly available to promote reproduction by other researchers. Reproducibility is validated through experiments at two separate laboratories that independently constructed the robot. The resulting platform supports autonomous control and teleoperation, serving as a suitable testbed for future investigations into data-driven navigation solutions.

\section{Related Work}

Tensegrity robotics research has demonstrated diverse abilities, including rolling~\cite{vespignani2018design}, crawling~\cite{kobayashi2022soft}, vibrating~\cite{rieffel2018adaptive}, climbing~\cite{friesen2014ductt}, and flying~\cite{mintchev2018soft}. While many studies are constrained to laboratory settings, some have demonstrated locomotion on unstructured terrain~\cite{wang2019light} or field environments~\cite{chen2017soft}. Various tensegrity simulators have been developed, those solving analytical differential equations~\cite{tmo, stedy, motes}, those ~\cite{Paul2006DesignAC, ntrt, caliper} building upon general-purpose physics engines, and, more recently, learning-based simulators~\cite{kun_sim2sim, kun_recurrent, kun_r2s2r, Chen2024LearningDT} to address the simulation-to-reality (sim2real) gap. Typical solutions for path planning and navigation involve geometric solutions that tile the robot's base polygons from start to goal~\cite{vespignani2018steerable,baines2020rolling}, often ignoring dynamic effects. Furthermore, real hardware experiments have frequently been limited to open-loop control~\cite{chang2018path,feng2024trajectory}. 

In contrast, the approach presented here uses a physics engine to accurately model motion and utilizes the robot's current pose as feedback. This re-planning compensates for unmodeled variables, enabling effective navigation through environmental disturbances, such as vertical drops, inclines, and granular terrain, and among obstacles.

\section{Robot Design}


\begin{figure}[tb]
    \centering
    \includegraphics[width=\linewidth]{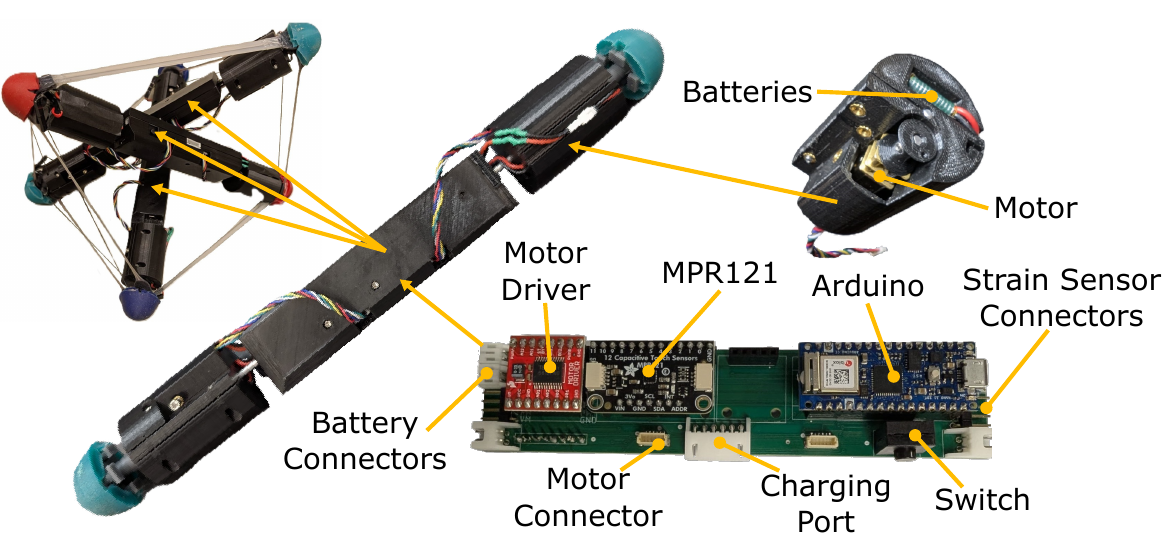}
    \vspace{-.3in}
    \caption{Open-source design: Each strut has 2 motor enclosures; each housing a brushed DC motor and LiPo batteries.  A custom motherboard uses commercial-off-the-shelf circuits mounted on headers, allowing for easy replacement.  A power switch and charging port provide convenience. The tensegrity robot is formed by assembling 3 struts with strain sensors.}
    \label{fig:design}
    \vspace{-0.25in}
\end{figure}

This work utilizes a 3-bar tensegrity robot, designed to be lightweight, deformable, and impact-resistant (Fig.~\ref{fig:splash}). To demonstrate the generality of the proposed navigation system, experiments are conducted using 3 distinct versions: an initial design (described in previous work), an upgraded version produced at a second laboratory, and a new open-source version developed for broad community adoption. The design files and assembly instructions for the latter two versions are available at \href{https://sites.google.com/view/tensegrity-navigation/}{https://sites.google.com/view/tensegrity-navigation/}.

The open-source design aims to facilitate easy reproduction and adoption by the robotics community. To this end, the new design features several key upgrades: \begin{itemize} 
\item A custom PCB enabling solderless (re)assembly. 
\item Convenient ports for peripherals and battery charging. 
\item WiFi communication via the Arduino Nano 33 IoT. 
\item A modular design where each bar contains a full electronics set, allowing the struts to be reconfigured for different tensegrity topologies. 
\item Open-source code and CAD files for user customization. 
\end{itemize}

{\bf Mechanical Design} The robot's 3 bars are structurally identical, differentiated only by colored end caps for pose tracking. Each bar houses 2 motors with winches to control tendon length. The robot features 9 tendons: 6 are actuated by the motors, and 3 function as passive restoring springs. All tendons incorporate capacitive strain sensors~\cite{johnson2022sensor}. On actuated tendons, the sensor runs in parallel with the motor's cable. On passive tendons, a more substantial sensor element is used, which dually serves as the restoring spring. These sensors enable closed-loop control of tendon lengths and assist in pose estimation. As shown in Fig.~\ref{fig:design}, the struts are assembled from 3D-printed enclosures and commercial off-the-shelf (COTS) parts. Each strut contains 2 motor sub-assemblies (housing motors and batteries) and a central electronics enclosure. The full robot is formed by connecting the strut assemblies with the tensegrity tendons.

{\bf Electrical Design} Each bar contains an Arduino Nano 33 IoT, which communicates with an off-board base station via WiFi. The Arduino is mounted on a custom printed circuit board (PCB) that also integrates a motor driver (TB6612FNG; SparkFun) and a capacitive sensing (MPR121; Adafruit) breakout board. This custom PCB includes a pushbutton switch, a 2 $\times$ 2S LiPo charging port, and auxiliary ports for future expansion. The compact electronics enclosure is designed to provide easy access to the switch, charging port, and the Arduino's microUSB port for programming. The electrical design is open-source and leverages COTS breakout boards to simplify assembly and maintenance.

{\bf Software Design} The open-source software comprises a ROS package for control and the on-board Arduino code. The primary ROS control node receives strain sensor data from each Arduino to implement a low-level PID controller. This node sends a control signal to each motor based on the error between the tendon's measured length and its target length. The Arduino translates this signal into a PWM direction and speed command for the motor. A "target shape" is defined by a set of 6 target lengths, 1 for each actuated tendon. A "motion primitive" is a sequence of these target shapes that constitutes a higher-level behavior. The following sections detail the other open-source components of the navigation solution: pose estimation, modeling motion primitives, and the use of these models for path planning and navigation.

\begin{figure*}[tb]
    \centering
    \includegraphics[width=0.95\linewidth]{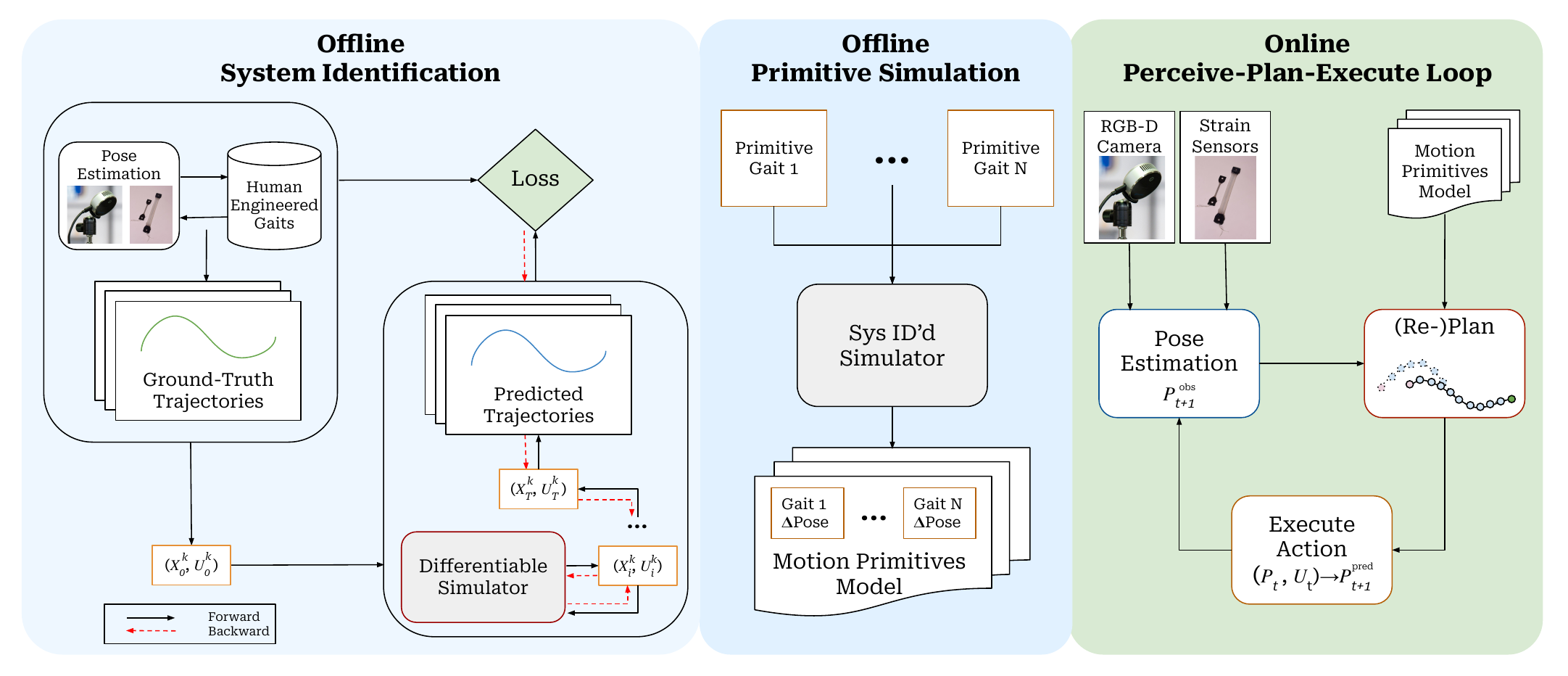}
    \vspace{-.2in}
    \caption{(Left) Offline sysID: Real trajectories are generated by chaining  human-engineered gaits and tracking the robot's pose via vision. Then, the start state $\mathbf{X}_0$ and controls $\mathbf{U}_{0:T-1}$ are autoregressively inputted into a differentiable simulator to predict corresponding trajectories. A scalar loss function between the ground truth and predicted trajectories is back-propagated to update system parameters. (Middle) Offline primitive modeling: Multiple human-engineered and in-simulation-found gaits are executed in the identified simulator. The changes in the pose of each primitive are recorded and stored in a motion primitive library. (Right) Online loop: As the tensegrity moves, the pose is  estimated at the start and end of each primitive motion. Then a plan from the tensegrity's current pose to the goal is computed, and its first primitive action executed. This is repeated until the tensegrity reaches the goal.}
    \vspace{-.2in}
    \label{fig:pipeline}
\end{figure*}

\section{Pose Estimation}
\label{section: pose estimatation}

The tensegrity robot's pose is tracked using a previously developed perception algorithm~\cite{tensegrity_perception}, which fuses data from an overhead camera (Intel RealSense L515) and on-board strain sensors. This algorithm operates in 2 steps. A local pose transformation is computed by scan-matching registered model points to observed points that fall within a maximum distance threshold and matching HSV color ranges. Then, this local transformation is jointly optimized with strain sensor measurements and known physical constraints. This process outputs the endpoints for each rod, which are used to infer the rod's pose. As this process resolves, however, only the rod's center axis (via its endpoints), the rod's twist information is lost due to its radial symmetry.

\section{Modeling Motion Primitives}
\label{sec:modeling}

The navigation solution requires the SE(2) transformations from the robot's motion primitives as input. As large-scale data collection on the physical robot is impractical, the system relies on simulation to obtain accurate predictions for each motion primitive. As shown in Fig.~\ref{fig:pipeline}, this offline modeling process involves system identification (sysID) and primitive simulation.


An accurate simulator requires identifying system parameters (sysID) that best match observed data. This work leverages previous differentiable tensegrity simulators \cite{kun_sim2sim, kun_recurrent, kun_r2s2r} to perform sysID via gradient-based optimization.

\noindent {\bf First-principles simulator} In \cite{kun_sim2sim}, a differentiable simulator $\mathcal{F}(\cdot)$ was developed, parameterized by $\Theta$, with input  the current state and controls $(\mathbf{X}_t, \mathbf{U}_t)$, and predicts the next state $\mathbf{\hat{X}}_{t+1}$ using Newtonian physics principles and an impulse-based linear contact model: \vspace{-.2in}

\begin{align}
    \mathbf{X_{t+1}}=\mathcal{F}(\mathbf{X_t}, \mathbf{U_t};\Theta)
    \vspace{-.15in}
\end{align}

$\Theta$ is optimized via gradient descent over a loss function $\mathcal{L}(\mathbf{X}_{t+1}, \mathbf{\hat{X}}_{t+1})$ between the predicted and observed output: \vspace{-.35in}

\begin{align}
    \Theta \leftarrow \Theta -\alpha\nabla_{\Theta}\mathcal{L}
        \vspace{-.3in}
\end{align}

\noindent {\bf System parameters} $\Theta$ are the contact parameters:
\begin{itemize}
    \item $\mu$ coefficient of friction between end caps and ground.
    \item $\epsilon$ coefficient of restitution.
    \item $\beta$ Baumgarte stabilization coefficient.
\end{itemize} 

\noindent Optionally, the cable stiffnesses $K$ and damping coefficients $\sigma$ can also be optimized in this process. In practice, we simply set the cables to have a high stiffness of $10^6 N/m$.

\noindent {\bf Training process}  First, a few trajectories $(\mathcal{T}^1,...,\mathcal{T}^k)$, where $\mathcal{T}^i:=[(\mathbf{X}^i_0, \mathbf{U}^i_0),...,(\mathbf{X}^i_{N-1},\mathbf{U}^i_{N-1}),(\mathbf{X}^i_N)]$, using human-engineered gaits, are collected. The robot's pose is tracked using the pose estimation algorithm of Section \ref{section: pose estimatation}. Next, corresponding predicted trajectories $(\mathcal{\hat{T}}^1,...,\mathcal{\hat{T}}^k)$ are generated by taking the ground-truth start state and applying $\mathcal{F}$ auto-regressively $N$ times: \vspace{-.2in}

\begin{align}
    \mathbf{\hat{X}}_N=\mathcal{F}^N(\mathbf{X}_0, \mathbf{U}_{0:N-1})
\vspace{-0.3in}
\end{align}

\begin{figure}[tb]
    \centering
    \includegraphics[width=\linewidth, angle=180]{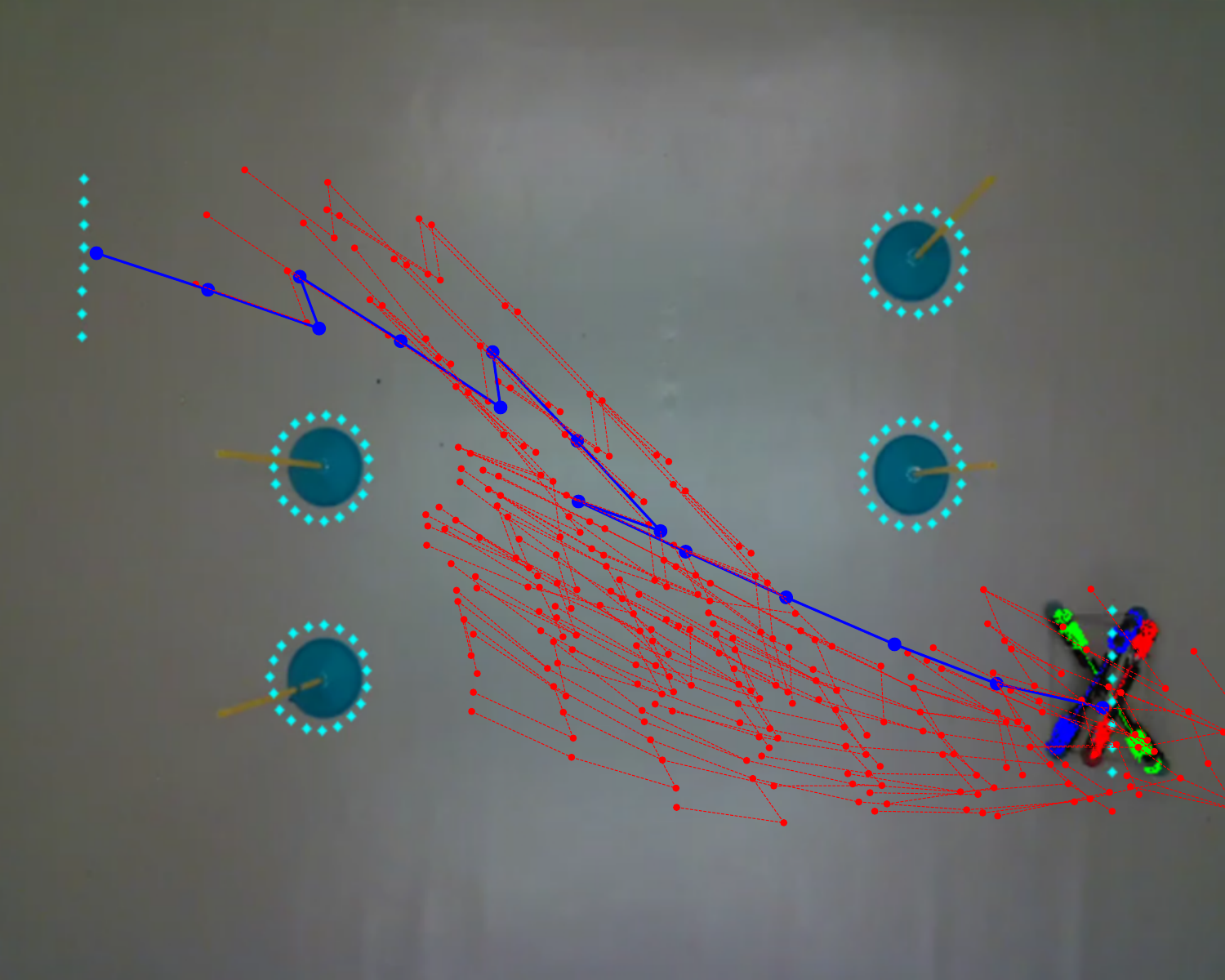}
    \vspace{-.05in}
    \caption{Example open-loop plan (dark blue) around obstacles (light blue circles) with unused expanded configurations (red).}
    \vspace{-.15in}
    \label{fig:planning}
\end{figure}

\noindent The system parameters are then optimized over the training data using gradient descent on the mean-squared error (MSE) loss between predicted and observed end cap positions. \vspace{-.2in}

\begin{align}
    \mathcal{L}_m(\mathbf{X}_m,\mathbf{\hat{X}}_m)&=\lvert\rvert \mathbf{X}_m - \mathbf{\hat{X}}_m\lvert\rvert^2_2 \\
    \mathcal{L}(\mathcal{T},\mathcal{\hat{T}})=\frac{1}{M}\sum_{m=0}^M{\mathcal{L}_m}&=\frac{1}{M}\sum_{m=0}^M\lvert\rvert \mathbf{X}_m - \mathbf{\hat{X}}_m\lvert\rvert^2_2 
    \vspace{-.2in}
\end{align}

Addressing partial observability (Section \ref{section: pose estimatation}), each forward pass rolls out the entire trajectory ($M$ gait cycles) and measures the pose mean squared error (MSE) at the end of each cycle, after which gradients are back-propagated. The converged system parameters are then validated by comparing a new real-world trajectory against its simulated counterpart.

\begin{table}[h]
    \centering
    \begin{tabular}{|c|c|c|c|}
         \hline
         Primitive & Gait & Left Max (mm) & Right Max (mm) \\
         \hline
         0 & Forward Roll & 200 & 200 \\
         \hline
         1 & Forward Roll & 220 & 220 \\
         \hline
         2 & Forward Roll & 240 & 240 \\
         \hline
         3 & Forward Roll & 200 & 220 \\
         \hline
         4 & Forward Roll & 220 & 200 \\
         \hline
         5 & Forward Roll & 200 & 240 \\
         \hline
         6 & Forward Roll & 240 & 200 \\
         \hline
         7 & Forward Roll & 220 & 240 \\
         \hline
         8 & Forward Roll & 240 & 220 \\
         \hline
         9 & Counterclockwise & 200 & 200 \\
         \hline
         10 & Clockwise & 200 & 200 \\
         \hline
    \end{tabular}
    \vspace{-0.1in}
    \caption{\vspace{-.05in}Parameters for the 11 motion primitives}
    \vspace{-0.35in}
    \label{tab:primitives}
\end{table}

This training process is iterated until the desired simulation-to-reality accuracy is achieved. Once validated, motion primitives can be reliably simulated and compiled into a library for planning. In practice, convergence is typically achieved in 2 iterations.

To ensure path planning tractability, a motion primitive library is constructed from human-engineered and simulation-generated gaits. This library contains 11 motion primitives (Table~\ref{tab:primitives}) classified into 3 types: (i) roll forward, (ii) turn clockwise, and (iii) turn counterclockwise. Gradual turning gaits are generated by modifying the roll-forward primitive. Specifically, the maximum tendon lengths on each side of the robot are varied, using values of 200~mm, 220~mm, and 240~mm to create nine distinct turning variations.

\section{Tensegrity Path Planning}
\label{sec:planning}

Discretizing the robot's possible actions into motion primitives reduces the continuous control problem to a discrete graph search. Each primitive is treated as an edge in a graph where nodes represent the robot's configurations in SE(2). This abstraction simplifies planning by focusing on achievable transitions between configurations rather than the detailed, low-level control inputs required to execute them. \vspace{-0.1in}

\begin{algorithm}[h]
    \caption{A* with Tensegrity Primitives}
    \begin{algorithmic}[1]
        \State \textbf{Input:} Start $\mathbf{S}$, Goal $\mathbf{G}$, Boundary $\mathbf{B}$, Obstacles $o$, Primitives $\mathbf{P}$
        \State \textbf{Output:} Path composed of primitives from $\mathbf{S}$ to $\mathbf{G}$
        \State Initialize Open List $\mathcal{O} = \{\mathbf{S}\}$ and Closed List $\mathcal{C} = \varnothing$ 
        \State Initialize $G_{score}[\mathbf{S}] = 0$
        \State Initialize heuristics $h$ by calculating distances along grid
        \While{$\mathcal{O}$ is not empty}
            \State $curr \gets$ pop node in $\mathcal{O}$ with the lowest $F_{score}[curr]$
            \If{\textbf{collisionDetect}($curr$, $\mathbf{B}$, $o$)}
                \State \textbf{continue}
            \EndIf
            \If{$curr$ within threshold of $\mathbf{G}$}
                \State \Return ReconstructPath($curr$, $cameFrom$)
            \EndIf
            \If{$curr$ within threshold of closest element in $\mathcal{C}$}
                \State \textbf{continue}
            \EndIf
            \For{each $p \in \mathbf{P}$}
                \State Child $child = curr$.propagate($p$)
                \State $G_{score}^\prime = G_{score}[curr] + \text{cost}(p)$
                \If{$G_{score}^\prime < G_{score}[n]$}
                    \State $cameFrom[child] = curr$
                    \State $G_{score}[child] = G_{score}^\prime$
                    \State $F_{score}[child]=G_{score}[child] + h(child)$
                    \State Add $child$ to $\mathcal{O}$
                \EndIf
            \EndFor
        \EndWhile
        \State \Return \textbf{Failure} (no path found)
    \end{algorithmic}
\end{algorithm}

\vspace{-0.2in}
\subsection{Open Loop Planner}
\label{subsection: open loop}

The open-loop planner employs a modified A* algorithm (Algorithm~1). In traditional A*, a closed list prevents revisiting identical states. In this application, however, the combination of motion primitives rarely generates the exact same pose twice. Instead, the search expands numerous similar but non-identical poses, leading to computational redundancy. To mitigate this, the planner uses a KD-Tree to efficiently query the nearest node in the closed list. This strategy allows the algorithm to prune new branches that are sufficiently close (within a predefined threshold) to an explored pose. As shown in Fig.~\ref{fig:planning}, this modification significantly reduces the number of expanded poses, enhancing planner efficiency at the expense of strict optimality. Despite this concession, the resulting planner is highly effective for real-time experiments where computational efficiency is important.

\subsection{Closed-loop Re-planner}
\label{sec:closed-loop-planner}

\begin{figure}[tb]
    \centering
    \includegraphics[width=\linewidth]{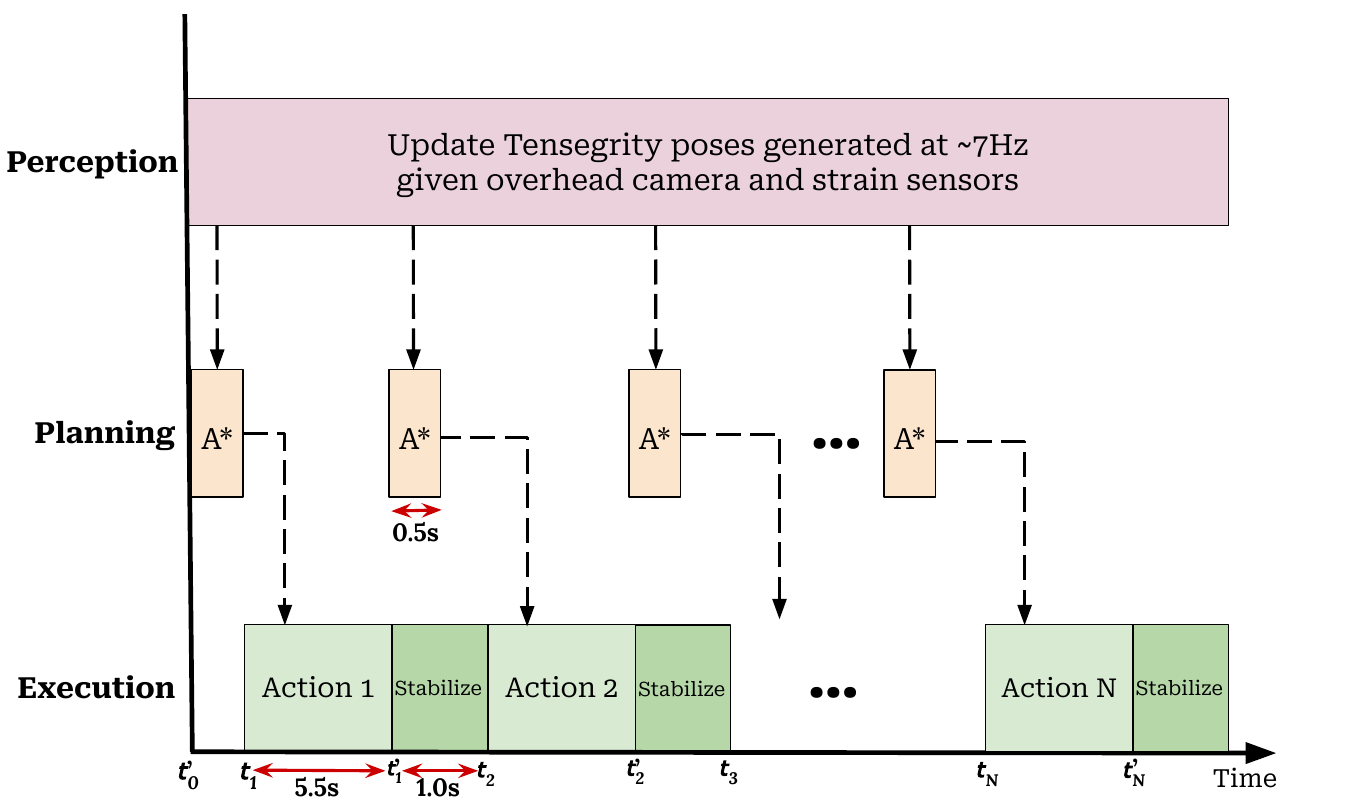}
    \vspace{-.25in}
    \caption{Online perception-plan-execution: Pose tracking runs at 7~Hz. The pose is passed to an A* planner. The first primitive in the new plan is executed. Re-planning occurs at the last step of each primitive as the robot returns to its rest state.}
    \vspace{-.25in}
    \label{fig:online workflow}
\end{figure}

Physical execution of motion primitives is subject to small variances, precluding perfect prediction of the robot's movements. Consequently, open-loop plans are effective only for short, undisturbed paths. Over longer trajectories, these small discrepancies between simulated and physical execution accumulate, resulting in pose error. To mitigate this, the system employs a closed-loop planner that generates a new path after each motion primitive is executed. The online workflow (Fig.~\ref{fig:online workflow}) involves continuous pose tracking at 7~Hz, with re-planning triggered after each primitive based on the current pose estimate. The planner may occasionally fail to find a valid path, often due to an inaccurate pose estimate or close proximity to an obstacle. In such cases (a "planning failure"), the robot executes the next primitive from the last valid plan. This fallback behavior continues until re-planning can be successfully performed.

\begin{figure}[h]
    \vspace{-.1in}
    \centering
    \includegraphics[width=\linewidth]{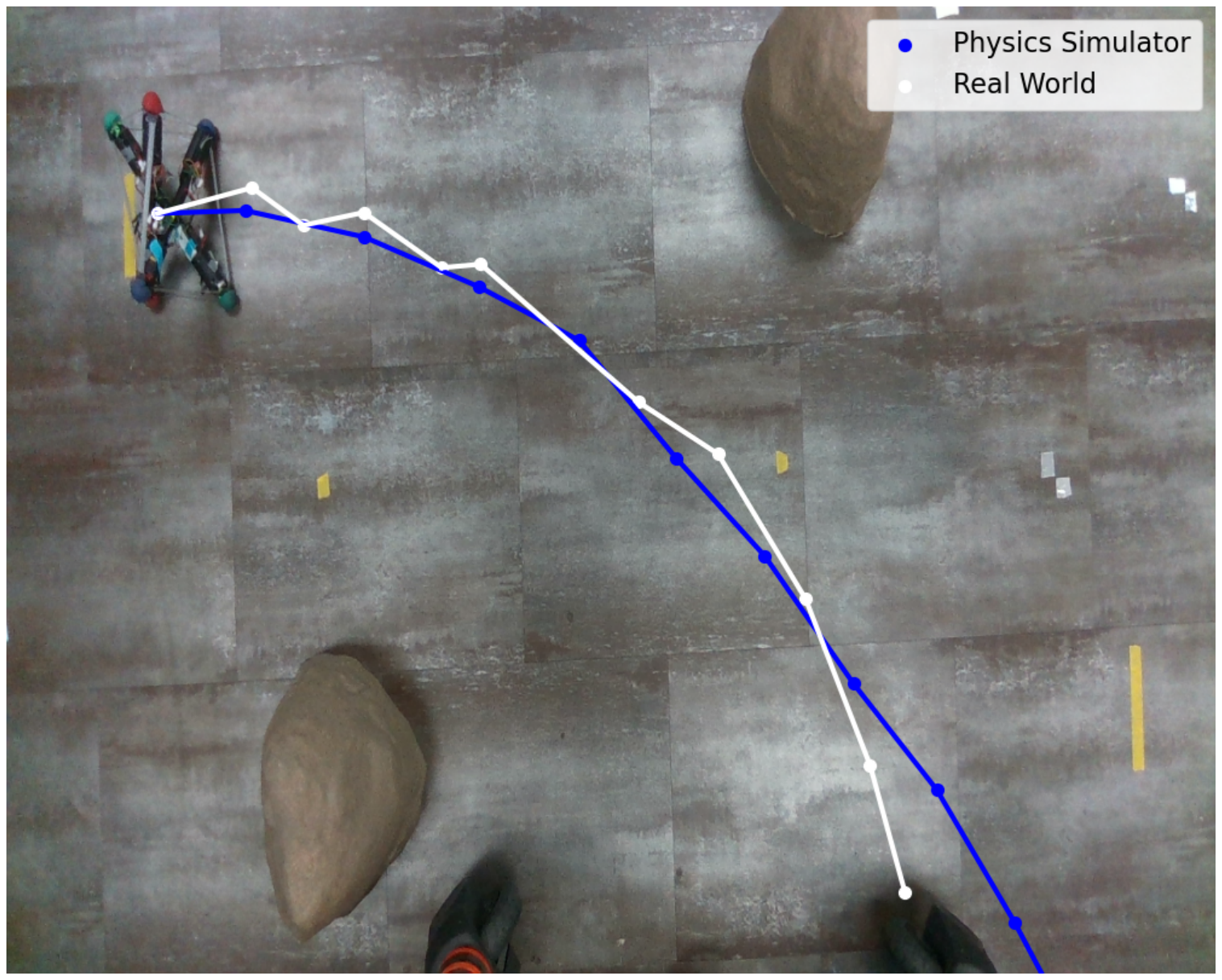}
            \vspace{-.35in}
    \caption{Comparison of an open-loop path in simulation (blue) and real (white) after sysID.  The simulation incorporates dynamic effects and accurately predicts the robot deviating from its path.}
 \vspace{-.1in}
    \label{fig:sim_real}
\end{figure}

\begin{figure}[tb]
    \centering
    \includegraphics[width=\linewidth]{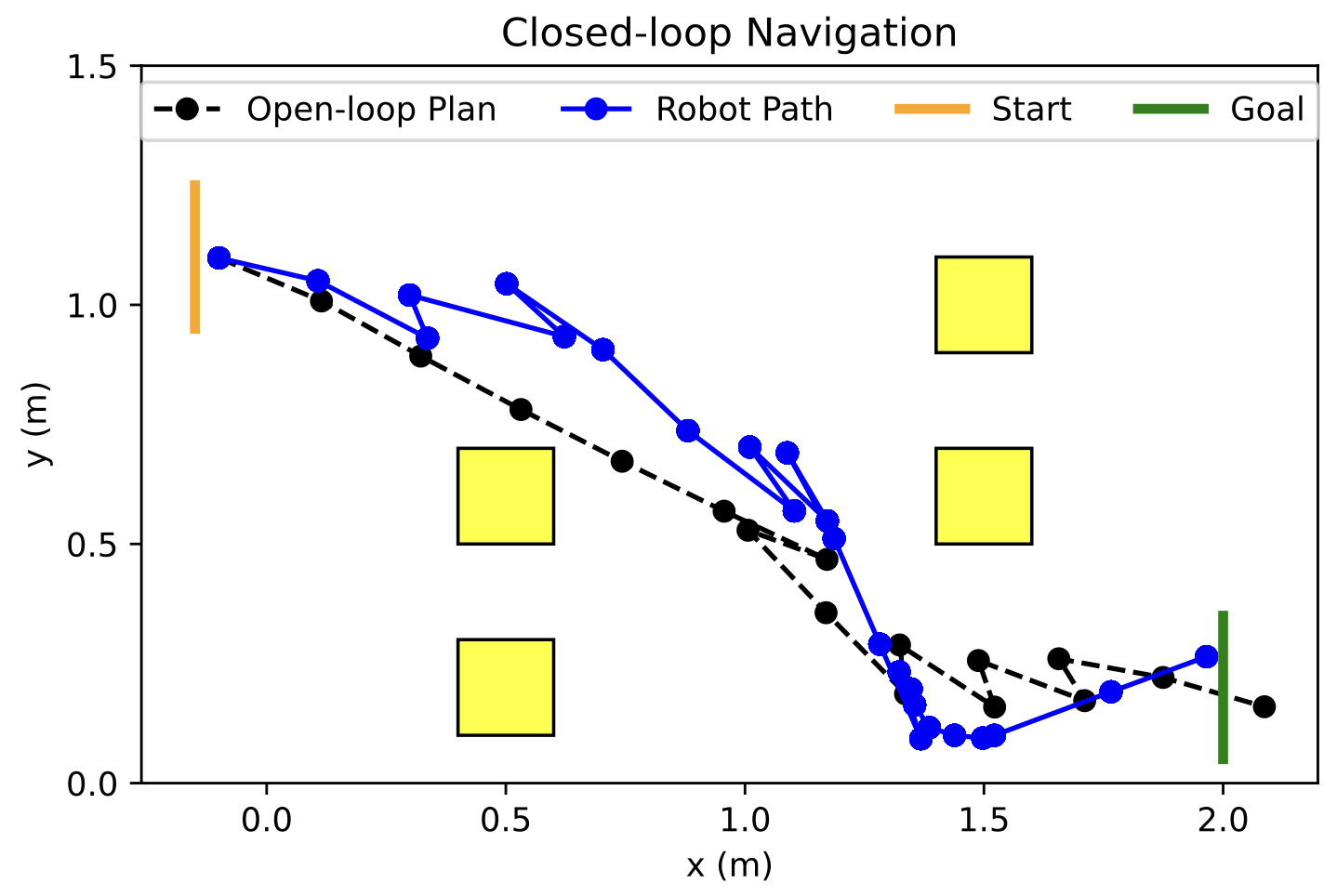}
    \includegraphics[width=\linewidth]{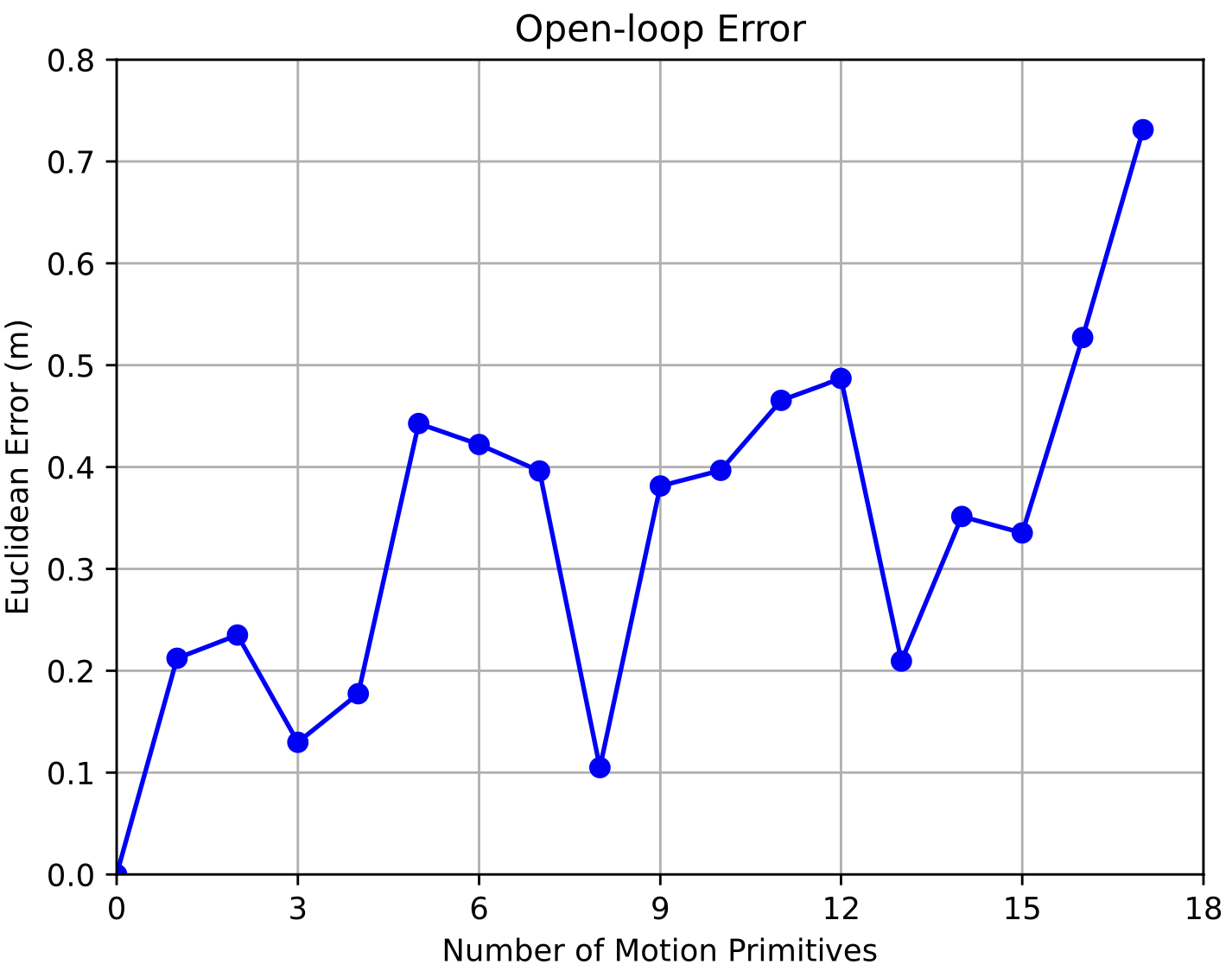}
    \vspace{-.3in}
    \caption{The path executed deviates from the plan, necessitating closed-loop control in the presence of obstacles.  (Top) The robot's path from an experiment against the open-loop path. (Bottom) The Euclidean distance between the open-loop plan and the robot's position as a function of the motion primitives executed.  As plans get more complex and require more steps, reliability of open-loop predictions decreases.}
    \vspace{-.25in}
    \label{fig:open_vs_closed}
\end{figure}

\section{Demonstrations and Experiments}

\begin{figure}[tb] 
    \centering
    \includegraphics[width=\linewidth]{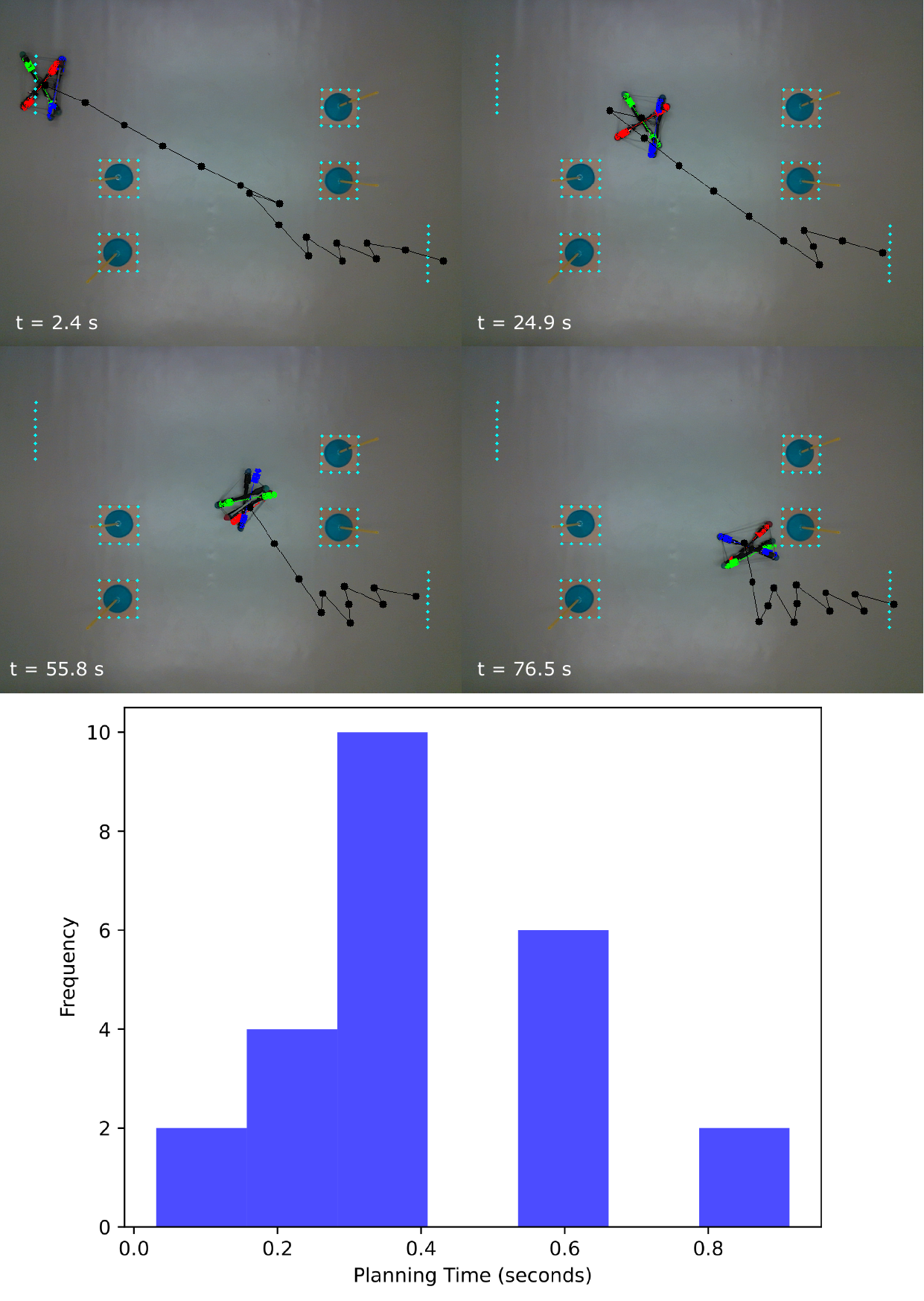}
        \vspace{-.25in}
    \caption{Navigating an obstacle course given an overhead camera.  The plan is recomputed after a motion primitive.  The histogram shows the distribution of planning compute times.  Planning time varies depending on where the robot is, i.e., how close to the obstacles and the goal.}
        \vspace{-.2in}
    \label{fig:closed-loop}
\end{figure}

\begin{figure}[tb]
    \centering
    \includegraphics[width=\linewidth]{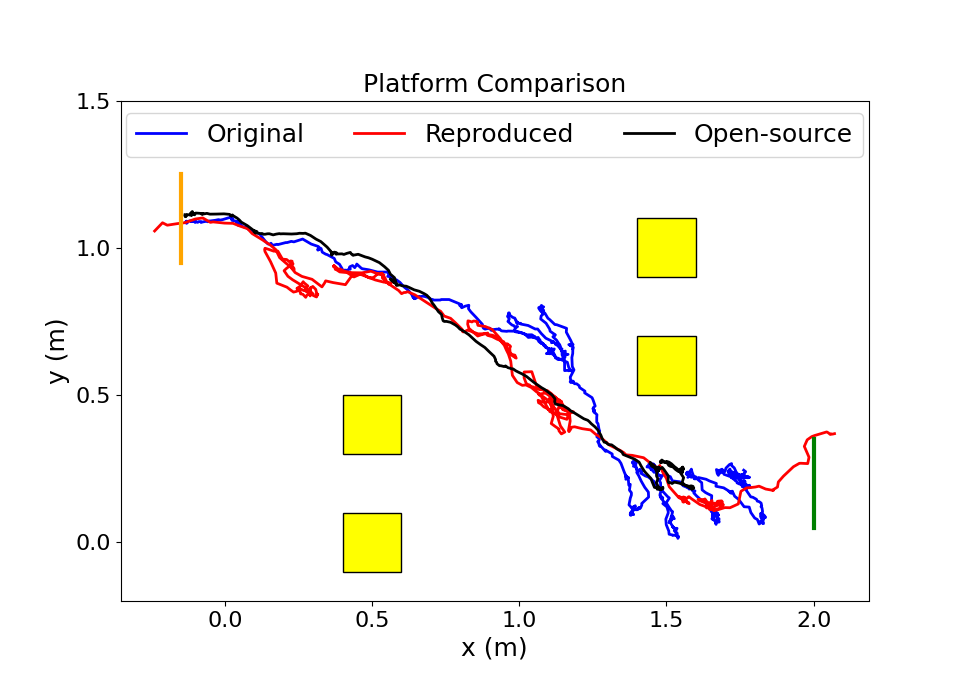}
    \vspace{-0.35in}
    \caption{Closed-loop executions by the 3 platforms given the same primitive model on the same obstacle course, i.e., without a re-identified model specific to each platform.  Re-planning addresses the model inaccuracies.}
    \vspace{-0.1in}
    \label{fig:platform_compare}
\end{figure}

\begin{figure}[tb]
    \centering
    \includegraphics[width=\linewidth]{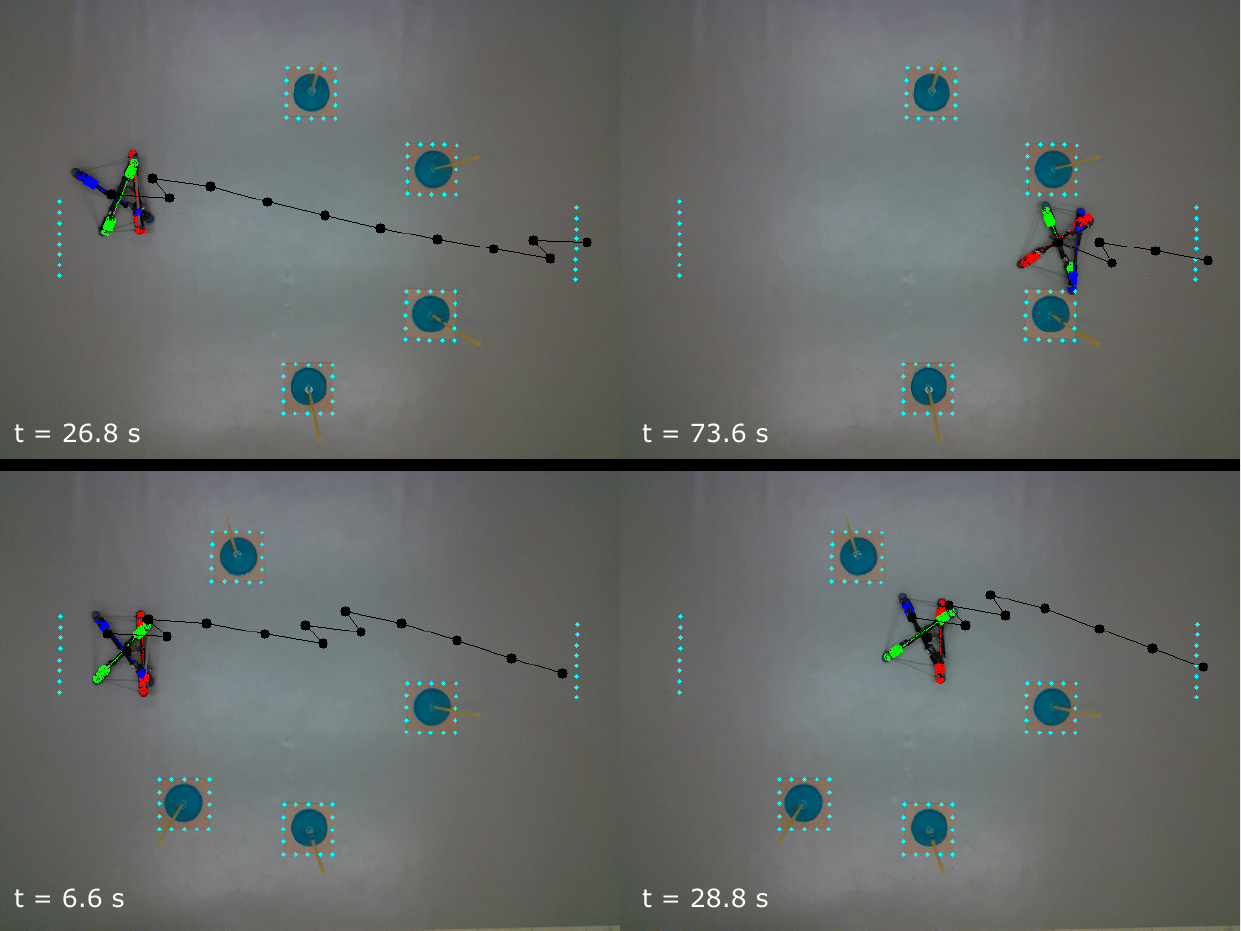}
    \vspace{-0.25in}
    \caption{Two different obstacle courses.}
    \vspace{-0.25in}
    \label{fig:diverse_scenes}
\end{figure}

Experiments demonstrate the robot's ability to autonomously navigate obstacle courses. An initial analysis of open-loop plan execution motivates the necessity of closed-loop re-planning for more complex fields. While many experiments are conducted in a standardized environment to compare the 3 platforms and evaluate reproducibility, the solution's effectiveness is also demonstrated in diverse settings. Furthermore, the closed-loop system's robustness is validated against unmodeled effects, including a vertical drop, an incline, and granular media. The system is also shown operating in an outdoor field. Videos of these experiments are available in the supplementary material.

{\bf Open-loop Execution} A 2D obstacle course was constructed to evaluate planner performance. Open-loop planning proved sufficient only for simple environments requiring few steps to reach the goal. As shown in Figure~\ref{fig:sim_real}, a robot executing an open-loop plan (post-sysID) successfully approaches the goal, but small, cumulative errors between the primitive models and the robot's physical dynamics cause it to deviate near the end of the path. This deviation highlights the necessity of a closed-loop re-planner.

{\bf Closed-loop Navigation} To mitigate open-loop limitations, the system employs the re-planner described in Section~\ref{sec:closed-loop-planner}. This re-planner adjusts the path after each primitive based on the robot's current pose, compensating for disturbances and enabling navigation through more challenging obstacle courses. Fig.~\ref{fig:open_vs_closed} compares the closed-loop path to the original open-loop plan, showing how the robot's executed path progressively deviates. As shown in Fig.~\ref{fig:closed-loop}, this closed-loop approach enables the robot to reliably reach the goal. Re-planning requires approximately 0.5~s on average, which is sufficient for real-time execution.

To demonstrate reproducibility, all 3 robot platforms (prior design, reproduced platform, and new open-source design) navigated an identical obstacle course (Fig.~\ref{fig:platform_compare}). Notably, all plans were derived from a motion primitive model based on the original platform's sysID. Despite lacking platform-specific models, all robots successfully completed closed-loop paths. This demonstrates that re-planning mitigates model inaccuracies, enabling robust navigation without needing to repeat sysID for each new robot. While many experiments were conducted in a standardized environment for comparative analysis, the closed-loop system also enables robust navigation in other obstacle fields (Fig.~\ref{fig:diverse_scenes}).

\begin{figure}[tb]
    \centering
    \includegraphics[width=\linewidth]{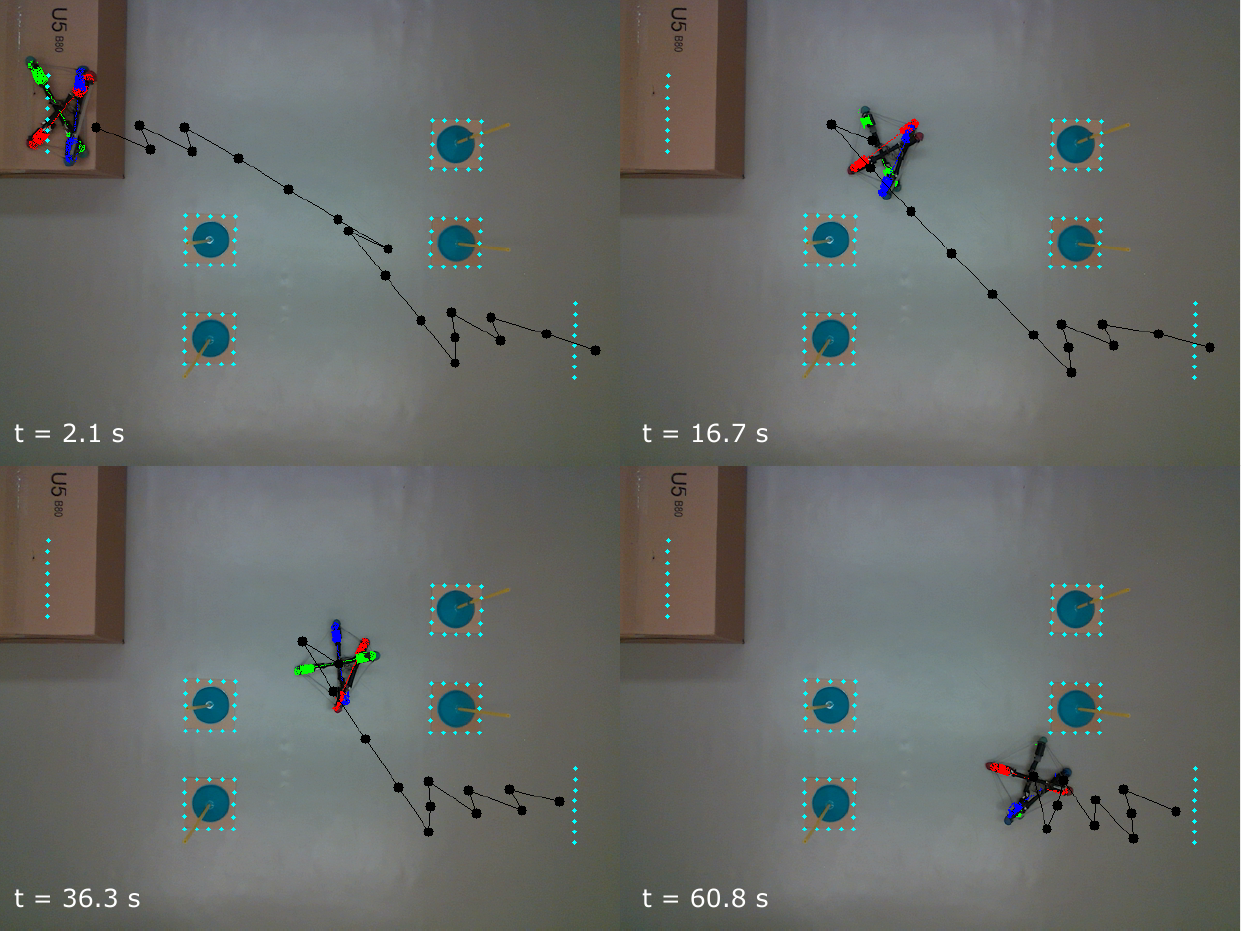}
    \vspace{-.3in}
    \caption{Feedback makes the navigation solution robust to disturbances, such as a 37~cm drop.  Due to the sudden movement from the drop, pose tracking lags but recovers.}
    \vspace{-.05in}
    \label{fig:drop}
\end{figure}

\begin{figure}[tb]
    \centering
    \includegraphics[width=\linewidth]{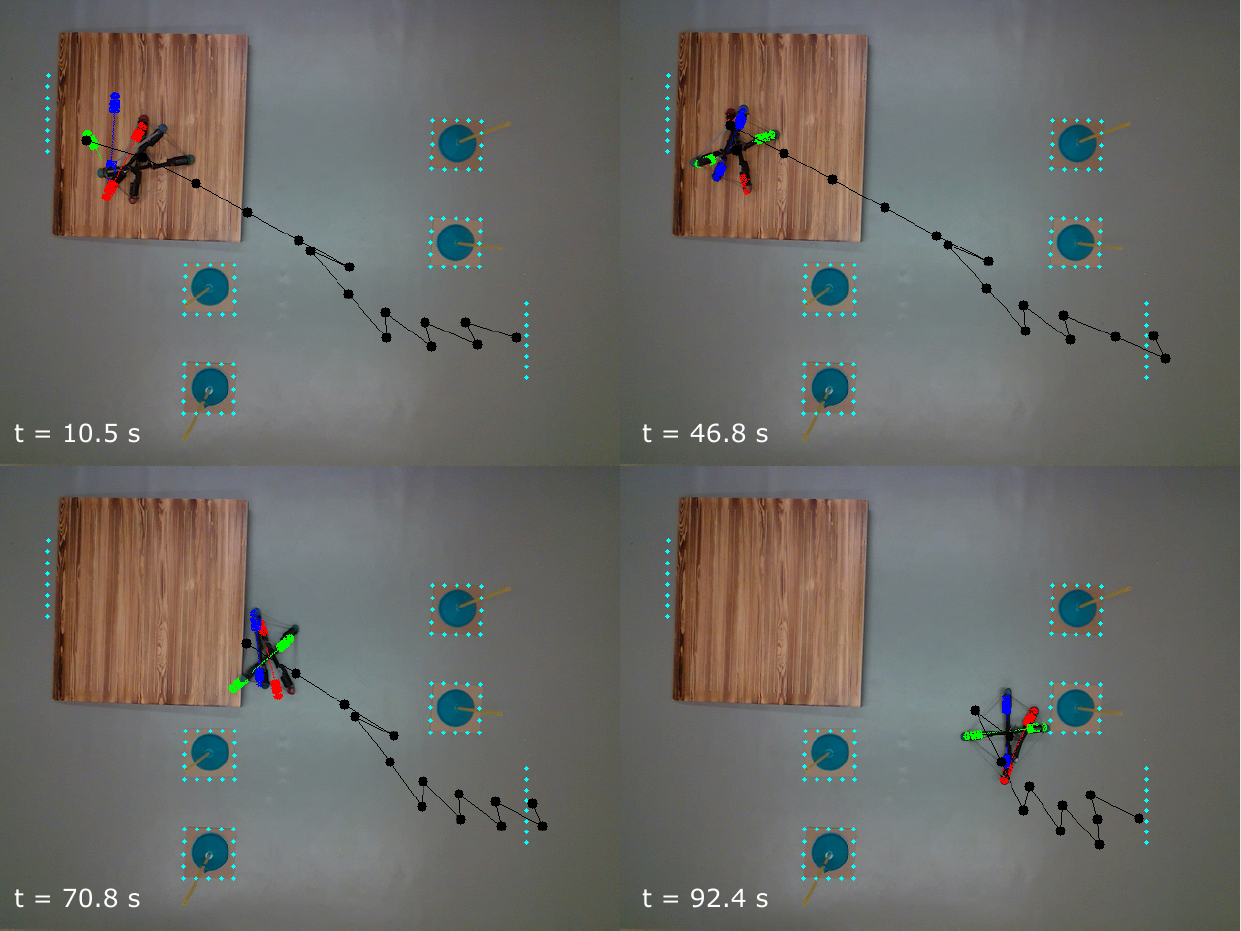}
    \vspace{-0.3in}
    \caption{The robot navigates an obstacle course with an 8\degree \ incline  wooden ramp without sysID on this new surface.  The approach is robust to errors in perception: toward the start, pose tracking gave inaccurate results, but eventually recovers.}
    \vspace{-0.2in}
    \label{fig:incline}
\end{figure}

\begin{figure}[tb]
    \centering
    \includegraphics[width=\linewidth]{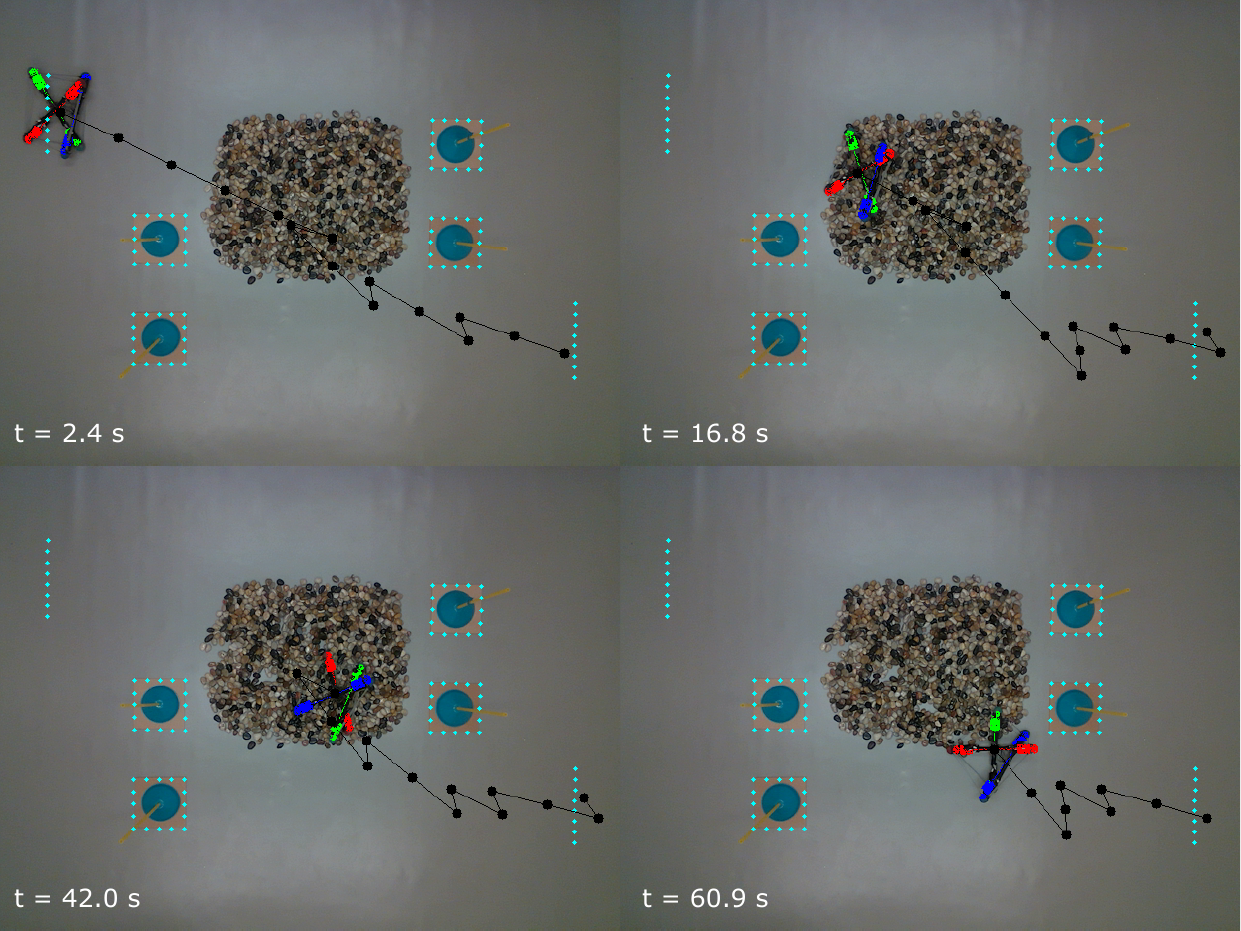}
    \vspace{-.25in}
    \caption{The solution is robust to unmodeled effects, such as granular terrain, where the primitives substantially differ from the flat ground for which sysID was performed.}
    \vspace{-.1in}
    \label{fig:granular}
\end{figure}

\begin{figure}[tb]
    \centering
    \includegraphics[width=\linewidth]{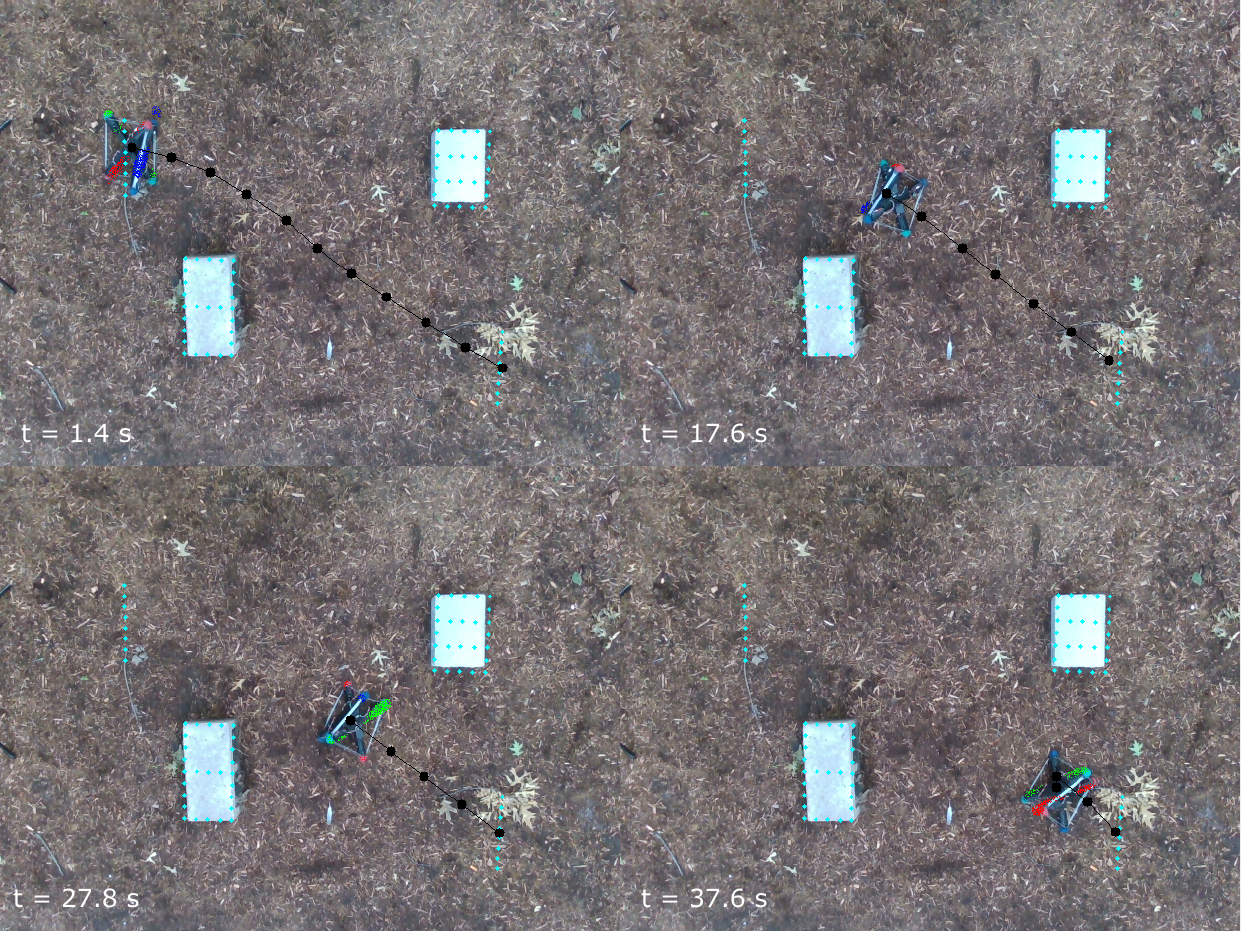}
    \vspace{-.25in}
    \caption{The tensegrity navigates an outdoor obstacle course.}
    \label{fig:field-test}
    \vspace{-0.2in}
\end{figure}

{\bf Drop} To evaluate the system's robustness to impacts, a strength of tensegrity robots~\cite{Johnson2025Impact}, a trial was conducted where the robot began at a height of 37~cm above the ground (Fig.~\ref{fig:drop}). Upon executing its first motion primitive, the robot rolled off the ledge, resulting in a post-drop pose significantly different from the model's prediction. The navigation pipeline proved robust to this disturbance. The pose tracker successfully estimated the robot's new position and orientation, allowing the re-planner to immediately calculate a new sequence of primitives from the post-drop state. Primitive gaits were modified via symmetry based on the robot's perceived orientation, enabling the robot to autonomously complete the obstacle course despite the unmodeled drop.

{\bf Incline} To further assess robustness, experiments were conducted on an 8\degree~wooden ramp (Fig.~\ref{fig:incline}). The incline functions as a  disturbance, as the motion primitives, modeled on flat ground, do not yield the same SE(2) transformations on a slope. Nevertheless, the solution successfully overcomes the incline without requiring repeated sysID. The system remains effective even as the pose tracking algorithm temporarily loses accuracy at the start of the trial, an issue caused by the ramp's distracting color.

{\bf Granular Terrain}  Repeating sysID for the vast range of terrains encountered in the field is infeasible. Therefore, the solution's robustness to unmodeled terrain effects was evaluated in an experiment requiring the robot to traverse granular media (Fig.~\ref{fig:granular}). The robot's primitive execution on granular media is significantly different from flat terrain, rendering the physics engine's predictions inaccurate. Despite this, the navigation system proved robust. The discrepancies between the model's prediction and the robot's physical execution were mitigated by the closed-loop planner, which generates a new path after each motion primitive.

{\bf Field Experiments} Field tests required switching from the LiDAR-based Intel RealSense L515 to the stereo-based D435, as the latter is operable in direct sunlight. This setup involved mounting the camera overhead, calibrating it with an April tag~\cite{apriltag}, and re-tuning the HSV filter values to track the robot's colored end caps under the new lighting conditions. Following these adjustments, the robot successfully navigated an outdoor obstacle course (Fig.~\ref{fig:field-test}). The closed-loop feedback system proved robust, compensating for the unmodeled effects of the outdoor environment.

\section{Conclusion}

This work presents a complete system for tensegrity robot navigation in environments with obstacles. The system models the robot's motion primitives using a physics engine and employs an A* planner to generate paths, incorporating feedback from a real-time pose tracking algorithm. The navigation pipeline demonstrates robustness to unmodeled environmental disturbances, including vertical drops, inclines, and granular media. Its effectiveness is also validated in an outdoor field environment. This contribution includes a complete open-source hardware design and software stack, enabling reproducibility of the system by other laboratories. This platform is intended to serve as a baseline for the robotics community to foster further advances in tensegrity navigation and related areas of robotics research.

\addtolength{\textheight}{-12cm}   






\bibliographystyle{IEEEtran}
\bibliography{references}

\end{document}